\documentclass[a4paper]{article}
\usepackage{ISCSLP2024}
\usepackage{ifthen}
\newboolean{blind}
\setboolean{blind}{false} 
\usepackage{tabularray}
\usepackage{hyperref}
\usepackage{csquotes}
\usepackage{multirow}
\usepackage{graphicx}
\usepackage{fbox}
\usepackage{tikz}
\usepackage[ruled,longend,linesnumbered]{algorithm2e}
\title{Improving Grapheme-to-Phoneme Conversion through In-Context Knowledge Retrieval with Large Language Models}
\name{
	\ifthenelse{\boolean{blind}}{Anonymous to ISCSLP}
	{Dongrui Han$^1$, Mingyu Cui$^2$, Jiawen Kang$^2$, Xixin Wu$^2$, Xunying Liu$^2$, Helen Meng$^2$}
}
\address{
  \ifthenelse{\boolean{blind}}{Anonymous to ISCSLP}
  {
  	$^1$Centre for Perceptual and Interactive Intelligence (CPII) Limited, Hong Kong SAR, China\\
  $^2$The Chinese University of Hong Kong, Hong Kong SAR, China
  }
}

\email{
	\ifthenelse{\boolean{blind}}{Anonymous to ISCSLP}
	{drhan@cpii.hk, mycui@se.cuhk.edu.hk, jwkang@se.cuhk.edu.hk, wuxx@se.cuhk.edu.hk, xyliu@se.cuhk.edu.hk, hmmeng@se.cuhk.edu.hk}
}

\begin{document}

\maketitle
\begin{abstract}
  Grapheme-to-phoneme (G2P) conversion is a crucial step in Text-to-Speech (TTS) systems, responsible for mapping grapheme to corresponding phonetic representations. However, it faces ambiguities problems where the same grapheme can represent multiple phonemes depending on contexts, posing a challenge for G2P conversion. Inspired by the remarkable success of Large Language Models (LLMs) in handling context-aware scenarios, contextual G2P conversion systems with LLMs' in-context knowledge retrieval (ICKR) capabilities are proposed to promote disambiguation capability. The efficacy of incorporating ICKR into G2P conversion systems is demonstrated thoroughly on the Librig2p dataset. In particular, the best contextual G2P conversion system using ICKR outperforms the baseline with weighted average phoneme error rate (PER) reductions of 2.0\% absolute (28.9\% relative). Using GPT-4 in the ICKR system can increase of 3.5\% absolute (3.8\% relative) on the Librig2p dataset.
\end{abstract}

\section{Introduction}

Text-to-Speech (TTS) synthesis aims to convert written text into speech with correct pronunciation and natural prosody.
Driven by the rapid development of deep learning, neural TTS systems such as \cite{ren2019fastspeech, wang2017tacotron, ren2020fastspeech, kim2021conditional} have been widely applied in practical human-machine
interactions.
To bridge the gap between linguistic text and acoustic pronunciation, Grapheme-to-Phoneme (G2P) serves as a front-end module in the current TTS systems to map written text to their corresponding pronunciation representations.
As any erroneous phonetic mapping could be propagated to the downstream modules and directly lead to mispronunciation, G2P plays an essential role in guaranteeing correct synthesis.

The initial works for G2P models adopt rule-based approaches, and later statistical models, such as Hidden Markov Models (HMMs), and Conditional Random Fields (CRFs) \cite{bellegarda2005unsupervised, taylor2005hidden, bisani2008joint,jiampojamarn2007applying} was used to model the relationship of grapheme and phoneme.
For the last decade, the development of deep learning led to the emergence of various neural G2P models, which leverage various network structures including Recurrent Neural Networks \cite{jyothi2017low}, Long-Short Term Memory \cite{rao2015grapheme, yao2015sequence} and convolutional networks \cite{yolchuyeva2019grapheme}.
Besides, transformer-based auto-regressive models have also been explored in G2P tasks \cite{yolchuyeva2020transformer, zhu2022byt5}.
More recently, SoundChoice \cite{ploujnikov2022soundchoice} was proposed as a sentence-level G2P model.
It employed homograph loss to resolve homograph disambiguation better, together with technologies such as curriculum learning and BERT embeddings \cite{devlin2019bert}, achieving promising performance on phoneme generation and homograph detection.


 \begin{table}[t]
\centering
\caption{Ambiguity example: The same written word "\textbf{wound}" has different pronunciations "\textbf{AW}" versus "\textbf{UW}" depending on different contextual meanings.}
\label{tab:homograph_example}
\resizebox{\columnwidth}{!}{%
\begin{tabular}{lll}
\hline
Sentence &
  Meaning &
  Phoneme \\ \hline
\begin{tabular}[c]{@{}l@{}}His string was \\ \textbf{wound} very tight\end{tabular} &
  \begin{tabular}[c]{@{}l@{}}describing the action\\ of  turning or twisting \\ something around\end{tabular} &
  W \textbf{AW} N D \\ \hline
\begin{tabular}[c]{@{}l@{}}Let me see the \\ \textbf{wound} on your leg\end{tabular} &
  \begin{tabular}[c]{@{}l@{}}An injury to living tissue\\  caused by a cut, blow, \\ or other impact\end{tabular} &
  W \textbf{UW} N D \\ \hline
\end{tabular}%
}
\vspace{-0.3cm}
\end{table}

    However, building accurate G2P datasets remains a laborious task, requiring extensive human annotation. Moreover, existing models often lack interpretability, making it difficult to understand the rationale behind certain phoneme outputs. Therefore, there is a need for more advanced techniques that can leverage contextual information effectively into the G2P conversion process.
    To this end, inspired by the remarkable success that Large Language Models (LLMs) like GPT-4 \cite{achiam2023gpt} demonstrated strong linguistic capabilities \cite{hewitt2019structural,bang2023multitask} and competence in related tasks like Automatic Speech Recognition (ASR) error correction \cite{mai2023enhancing}, contextual G2P conversion module with in-context knowledge retrieval from GPT-4 is proposed in this paper to provide richer and extensively semantic information in helping resolve the disambiguation gap in G2P mappings. The best-performing G2P conversion system using in-context knowledge retrieval outperforms the baseline without context information with phoneme error rate (PER) reductions of 2.0\% absolute (28.9\% relative) and homograph accuracy increase of 3.5\% absolute (3.8\% relative) on the Librig2p dataset.
    
\begin{figure}[t]
    \centering
    \includegraphics[trim=0 1mm 0 5mm, width=\linewidth]{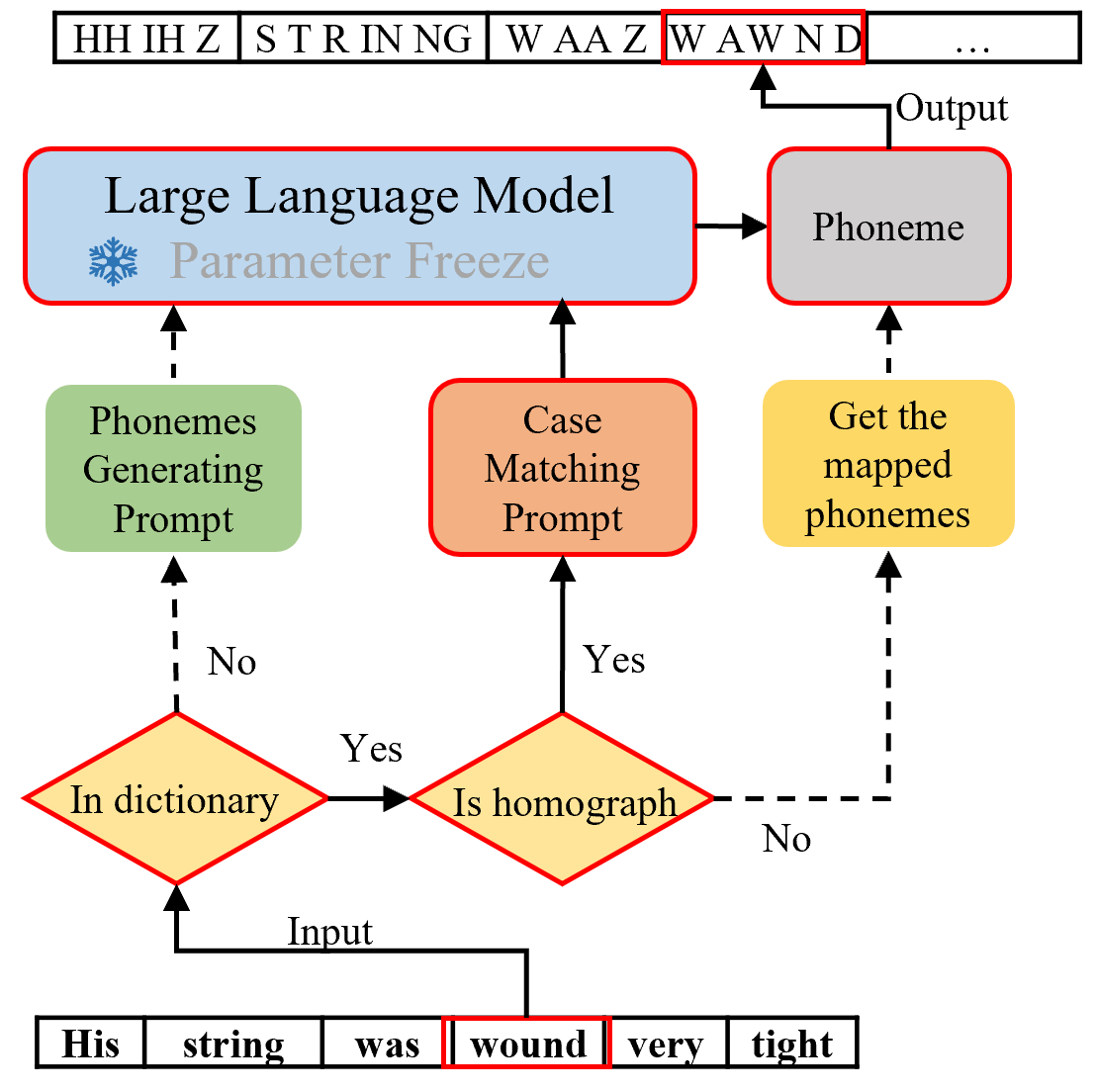} 
    \caption{LLM In-context Knowledge Retrieval of an example sentence: 1. Input word, and corresponding sentence to the system;  2. Tag each word if it is in the dictionary and is a homograph. 3. if the word is in the dictionary and is a homograph, then let LLM find the case where the usage of the word is closest to the context of the word input; 4 if the word is in the dictionary and is not homograph, then get the recorded phonemes directly; 5. If the word is not in the dictionary, then let LLM generate the phonemes of it considering the context.}
    \label{fig:look_up_dict}
\end{figure}

    The main contributions of this paper are summarized as follows:
    
    1) To the best of our knowledge, this is the first work that leverages in-context knowledge retrieval from GPT-4 for handling the disambiguation challenge in G2P mapping process for G2P systems. In contrast, previous research has been limited to data augmentation and architectural modifications, overlooking the potential of leveraging LLMs' contextual capabilities.

    2) The efficacy of utilizing in-context knowledge retrieval in G2P conversion systems is extensively shown on the Librig2p task and provides insights on their potential to benefit other non-context based G2P architectures.

    3) The highest accuracy rate of 95.7\% and the lowest weighted average phoneme error rate of 4.9 \% were obtained on Librig2p dataset compared to other non-context based methods, which demonstrates the potential of in-context knowledge retrieval for context-aware speech synthesis applications.

\section{Methods}
\label{section:method}
\vspace{-0.1cm}
\subsection{LLM One-shot Prompt}
\label{section:one-shot-g2p-prompt}
\vspace{-0.1cm}
GPT-4, developed by OpenAI \cite{achiam2023gpt}, is a powerful general-purpose language model with impressive capabilities across code generation, translation, math reasoning, and more \cite{mai2023enhancing}. We hypothesize that GPT-4's linguistic knowledge could facilitate the direct conversion of G2P by generating phoneme sequences from input sentences through a one-shot prompting approach.

\begin{figure}[b]
    \vspace{-0.4cm}
    \centering
    \includegraphics[trim=0 1mm 0 5mm, width=\linewidth]{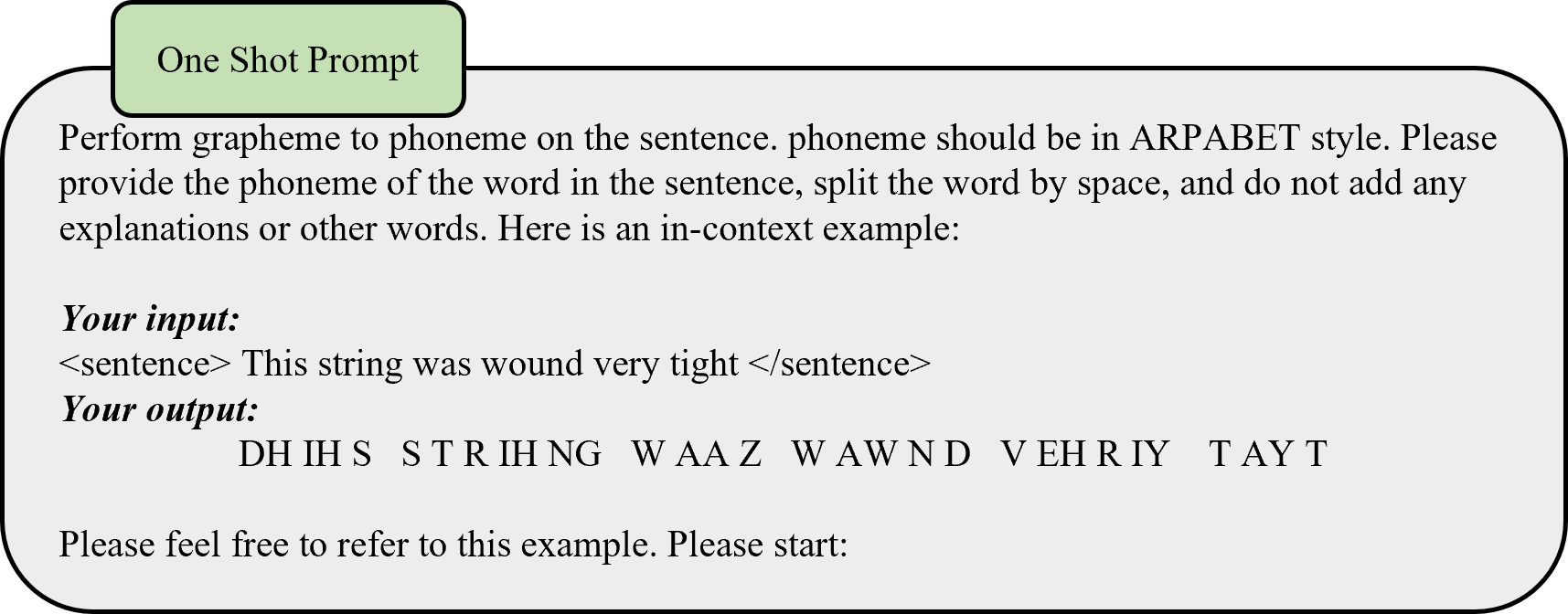}
    \caption{LLM One-shot prompt. Here is only the prompt without user input.}
    \label{fig:one-shot-prompt}
    \vspace{-0.4cm}
\end{figure}

Designing effective prompts is crucial for eliciting desired outputs from LLMs like GPT-4. The prompt should clearly specify the task at hand and outline the expected output format of the response to facilitate post-processing. Inspired by prior work \cite{ma2023can} on ASR error correction, we utilize the prompt template as \autoref{fig:one-shot-prompt}.

By providing a representative example mapping in the prompt, we aim to signal the desired input-output behavior to GPT-4. The model can then generate the phoneme sequence for the provided input sentence in a one-shot manner without any dataset-specific fine-tuning.
\vspace{-0.1cm}
\subsection{LLM In-context Knowledge Retrieval}
\label{section:look_up_dict}
\vspace{-0.1cm}
Our hypothesis is that GPT-4's extensive knowledge base contains broader linguistic context and semantics compared to specialized phoneme data. Consequently, GPT-4 will be better suited to extracting a word's meaning and part-of-speech from the sentence, rather than directly mapping to phoneme representations. Inspired by how humans look up word pronunciations in dictionaries, we propose an in-context knowledge retrieval based system that treats GPT-4 as an "AI linguist" for disambiguating homographs and generating phonemes. The proposed approach consists of the following steps:

\begin{enumerate}
    \item \textbf{Look up the word in a homograph}: The in-context knowledge retrieval dictionary is structured as a collection of entries, where each entry corresponds to a homographic word and contains multiple sub-entries, each representing a distinct meaning or part-of-speech of the word, along with its corresponding phoneme pronunciation, example usage context and definition.

    \item \textbf{Prompt GPT-4 to analyze and identify the input sentence}: The prompt provided to GPT-4 will include the target word and the complete sentence. The model will be instructed to output the most relevant meaning, part-of-speech, and contextual information for the target word based on the understanding of the sentence.

    \item \textbf{Retrieve the phoneme pronunciation}: The system will compare GPT-4's output with the sub-entries in the dictionary and select the sub-entry whose meaning, part-of-speech, and example context most closely align with GPT-4's analysis, retrieving the corresponding phoneme pronunciation.
    
\end{enumerate}
For non-homographic words with unambiguous pronunciations or words not found in the in-context knowledge retrieval, a separate special tag can be used for simple lookups. The key steps in this LLM In-context Knowledge Retrieval workflow are illustrated in \autoref{fig:look_up_dict}. GPT-4 plays the role of the linguist, leveraging its wide knowledge to disambiguate word senses and apply the appropriate dictionary pronunciation based on the sentence context.

This LLM In-context Knowledge Retrieval workflow leverages GPT-4's strengths in language understanding while constraining phoneme outputs to conform to curated dictionary entries, potentially improving performance over the unconstrained one-shot prompting method. The key steps of this workflow are summarized as follows:

\begin{figure}[t]
    \centering
    \vspace{-0.5cm}
    \includegraphics[trim=0 1mm 0 5mm, width=\linewidth]{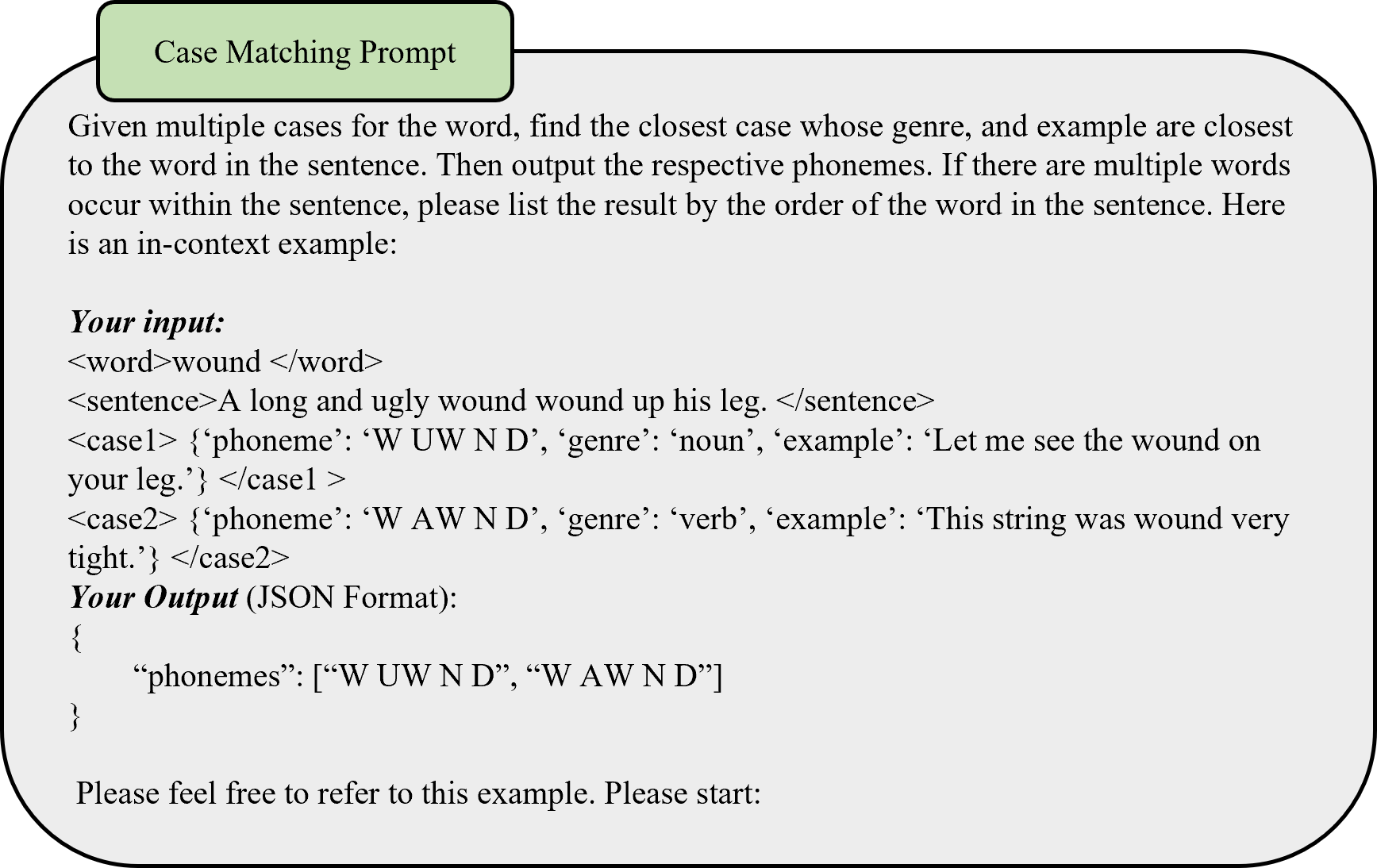}
    \caption{LLM case matching prompt. Here is only the prompt without user input.}
    \label{fig:case-matching-prompt}
    \vspace{-0.5cm}
\end{figure}

\textbf{Case matching:} This module involves matching the input word to the most relevant case in the dictionary based on the sentence context and then outputting the corresponding phoneme sequence. Since a single word can appear multiple times within a sentence, the prompt is structured as \autoref{fig:case-matching-prompt} to provide the target word and the sentence containing that word as inputs. The LLM will provide distinct phoneme outputs for the word based on its contextual meaning in each sentence.

\textbf{Sentence-level phoneme generation:} After obtaining the phoneme sequences for each word in the sentence, the system concatenates them in their original order, taking into account word boundaries, stress patterns, and any necessary post-processing to form the complete phonetic representation of the input sentence.

\begin{figure}[t]
    \centering
    \includegraphics[width=\linewidth]{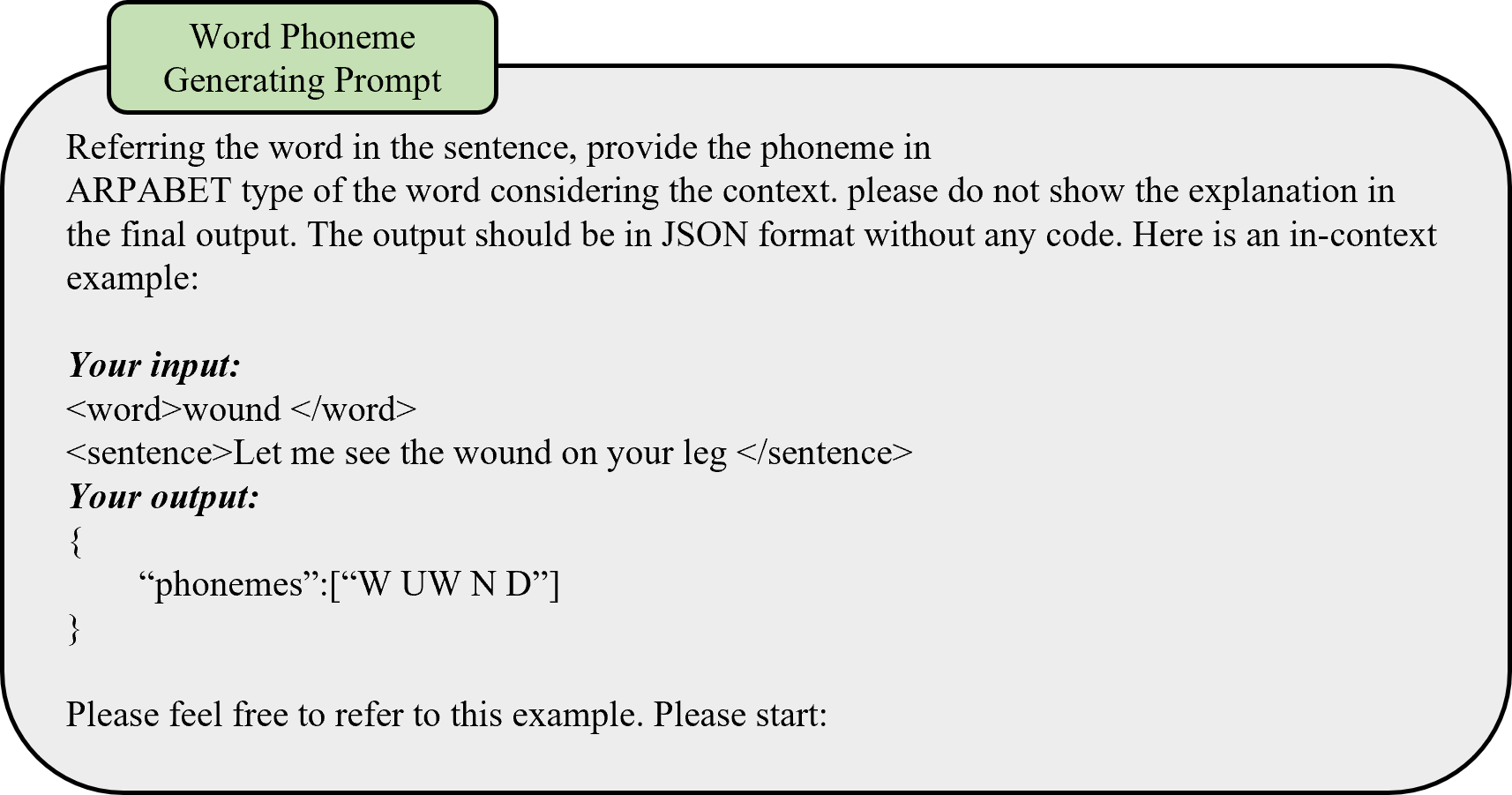}
    \vspace{-0.5cm}
    \caption{LLM word phoneme generating. Here is only the prompt without user input.}
    \label{fig:word-phoneme-gen}
\end{figure}

\section{Experiments}

\subsection{Experiments Setup}

Librig2p-nostress-space dataset proposed by SoundChoice \cite{ploujnikov2022soundchoice} is adopted to evaluate performance. Phonemes are initially presented as list objects in the dataset, we reformat them into strings to train and evaluate the PER easily. Additionally, since the dataset omits the stress makers of the phonemes, we drop the stress marker 0,1,2 from the generated phonemes.
To evaluate the performance of different models. This paper uses Phoneme Error Rate (PER)\footnote{\url{https://github.com/jitsi/jiwer}} and homograph accuracy to evaluate the performance of the models. The PER is obtained by calculating the Levenshtein Distance between the ground truth phoneme sequences and the generated phoneme sequences. The homograph accuracy is evaluated by calculating the accuracy of homographs.

The SoundChoice \cite{ploujnikov2022soundchoice} is trained on Librig2p-nostress-space dataset as the baseline system\footnote{\url{https://huggingface.co/speechbrain/soundchoice-g2p}}. The model is trained with 50 epochs on lexicon data, 35 epochs on sentence data, and 50 epochs on homograph data.

\textbf{LLM One-shot Prompt:} We notice that GPT-4 performs much better than GPT3.5 on word information extraction, such as extracting the genre, and definition of a word in a context. We choose the GPT-4 to gain the performance as better as possible. The GPT-4 version is \textit{GPT-4-0613}. We only set the role of \textit{\textbf{user}} when calling the GPT-4 API. Additionally, we fine-tune Llama2-7B-Chat and Gemma-2B-it on this task. The fine-tuned data are generated by reformatting the Librig2p dataset.

\textbf{LLM In-context Knowledge Retrieval:} The dictionary of in-context knowledge retrieval is generated by the Librig2p train dataset and the CMU dictionary [add ref]. The homographs and corresponding phonemes are obtained from the librig2p homograph train dataset. We only put the clearly defined homograph into the homograph dictionary, use GPT-4 to generate cases automatically, and modify the examples in case of any ambiguity. Each homograph contains multiple cases. Each case contains a corresponding \textit{\textbf{genre, phoneme, and example sentences}}. For the non-homograph words, besides adding words from the CMU dictionary, omit those with multiple phonemic transcriptions. we design two methods to add words from the Librig2p dataset as follows:
\begin{itemize}
    \item \textbf{librig2p\_omit}: From the Librig2p training dataset, include the words that only have a single phonemic transcription. Words with multiple phonemic transcriptions will not be considered.
    \item \textbf{librig2p\_freq}: Add words from the Librig2p training dataset. If a word has several phonemic transcriptions, only the transcription that appears most frequently within the dataset is included.
\end{itemize}
Considering the dictionary influence, we also change the leading position of the dictionary between the CMU dictionary, and the Librig2p data, for example, \textbf{cmu+Librig2p\_freq} versus \textbf{Librig2p\_freq+cmu}. Because many words are contained in both the CMU dictionary and Librig2p datasets, the phonemes for those duplicated words will get from the front one, such as \textit{cmu+librg2p\_omit}, if the word is duplicated in both data, then we will follow its phonemes in cmu.
Since there are no datasets to fine-tune LLama2-7B-chat, Gemma-2B-it for the \textbf{Case matching} prompt, we generate the example sentence using GPT-4. Then reformat them as the prompt example described in \autoref{section:look_up_dict}.

\vspace{-0.1cm}
\subsection{Fine-tuned Models}
\vspace{-0.1cm}

To get a customized and stable G2P system, using a local LLM to achieve an object is always an option. Fine-tuning LLM is a highly effective way to improve performance \cite{min2021metaicl,wei2021finetuned}. We chose Llama2-7B-chat\cite{touvron2023llama} and Gemma-2B-it\footnote{\url{https://huggingface.co/google/gemma-2b-it}} to work on the G2P task. 

We use QLoRA to fine-tune those models. QLoRA reduces the memory usage of LLM fine-tuning without performance trade-offs compared to standard 16-bit model fine-tuning. More specifically, QLoRA uses 4-bit quantization to compress a pre-trained language model. The LLM parameters are then frozen and a relatively small number of trainable parameters are added to the model as Low-Rank Adapters. During fine-tuning, QLoRA backpropagates gradients through the frozen 4-bit quantized pre-trained language model into the Low-Rank Adapters. The LoRA layers are the only parameters being updated during training. \cite{dettmers2024qlora}

\textbf{Word phonemes generating:} For out-of-vocabulary words not present in the dictionary, an additional prompt is used to have GPT-4 generate the phoneme sequence directly, while taking into account the word's context within the sentence. The prompt for this out-of-vocabulary generation step is structured as \autoref{fig:word-phoneme-gen}.

\vspace{-0.1cm}
\section{Results}
\subsection{Results on One-shot}
\vspace{-0.1cm}

The PER and homograph accuracy of LLM: one-shot method evaluated on the Librig2p test dataset are shown in \autoref{tab:result}. Several trends can be found:

1) The GPT-4 can applied for G2P conversion but performs poorly without customized fine-tuning. The PER has a significant reduction and is better than the baseline model after applying the fine-tuned model(Llama2-7B-chat, Gemma-2B-it) in G2P. The Fine-tuned Gemma-2B-it achieves the lowest weighted PER(5.2\%) among other models including the baseline model, which proves the capability of the LLM on G2P conversion.

2) The homograph accuracy is lower than the baseline model. Compared with the baseline model exploiting curriculum learning (that gradually switches from word-level to sentence-level G2P) with specified weighted homograph loss, the fine-tuned models are simply trained for 3 epochs. The homograph accuracy of the fine-tuned model is higher than GPT-4, which implies the model can learn the context information by fine-tuning to get a higher homograph accuracy.


\begin{table}[]
\centering
\caption{Comparing G2P between the baseline model and  LLM In-context Knowledge Retrieval}
\label{tab:compare_table}
\resizebox{\columnwidth}{!}{%
\begin{tabular}{ccc}
\hline
\multicolumn{1}{c}{Reference} &
  Baseline &
  \begin{tabular}[c]{@{}c@{}}LLM In-context \\ Knowledge Retrieval\end{tabular} \\ \hline
\begin{tabular}[c]{@{}l@{}}... THE \textbf{BOW} OF \\ VISHNU STRINGS... \end{tabular} &
  ...DH AH   \textit{B \textcolor{blue}{AW}} AH V ... &
  ...DH AH \textit{B \textcolor{red}{OW}} AH V ... \\ \hline
\begin{tabular}[c]{@{}l@{}} ... MAKING COMPLEX \\ \textbf{DIAGNOSES} SUCH AS ...\end{tabular} &
  \begin{tabular}[c]{@{}l@{}}... K AA M P L EH K S  \\ \textit{D AY AH G N OW S \textcolor{blue}{AH} Z}   \\ S AH CH...\end{tabular} &
  \begin{tabular}[c]{@{}l@{}}...K AA M P L EH K S\\ \textit{D AY AH G N OW S \textcolor{red}{IY} Z}   \\ S AH CH...\end{tabular} \\ \hline
\begin{tabular}[c]{@{}l@{}}... ONE BID \textbf{INCREMENT} \\ ABOVE THE ... \end{tabular} &
  \begin{tabular}[c]{@{}l@{}}...B IH D   \\ \textit{IH N K R AH M \textcolor{blue}{EH} N T} \\ AH B AH V...\end{tabular} &
  \begin{tabular}[c]{@{}l@{}}...B IH D   \\ \textit{IH N K R AH M \textcolor{red}{AH} N T} \\ AH B AH V...\end{tabular} \\ \hline
\end{tabular}%
}
\vspace{-0.5cm}
\end{table}

\vspace{-0.1cm}
\subsection{Results on LLM In-context Knowledge Retrieval}
\vspace{-0.1cm}

The PER and homograph accuracy of \textit{LLM In-context Knowledge Retrieval} system is shown in \autoref{tab:result} and \autoref{tab:addon_result}. Several trends can be found:

1) The weighted average PER is lower than the baseline model for each LLM, which verifies the feasibility of this method. The fine-tuned model achieves the lowest weighted average PER among others because it can generate the phonemes of OOV words following the distribution of the Librig2p dataset. 

2) The homograph accuracy of GPT-4 is the highest among other models. Since the training data of \textbf{case matching prompt} for this method is generated by GPT-4, it is logical that the homograph accuracy of the fine-tuned model is lower than GPT-4. The result implies that the strong linguistic capabilities of GPT-4 \cite{hewitt2019structural,bang2023multitask} can help identify the homograph in a sentence, and GPT-4 has much better semantic understanding capabilities than the other two fine-tuned models. The result proved our hypothesis in \autoref{section:look_up_dict} that \textit{``GPT-4's broad knowledge base covers more linguistic context and semantics than specialized phoneme data.''}.

3) For GPT-4, the LLM In-context Knowledge Retrieval system shows a significant improvement compared to the one-shot method. The result implies the feasibility of the system.  

4) Here we compare different ways of dictionary-building to check if the method can be applied by other LLMs. The result shows that fine-tuned models also achieve some relatively good performance. Meanwhile, the dictionary \textbf{librig2p\_freq} makes the method achieve a lower weighted average PER than the method of \textbf{``one-shot''}. Whether omit the words that have multiple phonetic representations in datasets has a big influence on the result. The distribution of the Librig2p train dataset is closer to the Librig2p test dataset, thus the Librig2p\_freq can lead to a better performance than the Librig2p\_omit. 

5) The result of the extra experiment \autoref{tab:addon_result} shows that the order of the dictionary barely influences the final result. Besides the loss of precision from dictionary building, the PER should also come from OOV words and homographs. Compared to previous work, this system is more extensional. As long as the semantic understanding capabilities of LLM are sufficient to distinguish the role of words in different contexts, a high-quality dictionary is a key point to promote the system. The dictionary can be customized by preference without model training. How to customize a high-quality phoneme dictionary is a challenge for this system in the future.

\begin{table}[]
\centering
\caption{Performance in terms of PER(\%) and Homograph Accuracy(\%) for all combinations of methods and models. The dictionary 1 represent cmu+librg2p\_omit, 2 represent cmu+librg2p\_freq}
\vspace{-0.3cm}
\label{tab:result}
\resizebox{\columnwidth}{!}{%
\begin{tabular}{ccccccccc}
\hline
\textbf{ID} &
  \textbf{method} &
  \textbf{model} &
  \textbf{dictionary} &
  \textbf{\begin{tabular}[c]{@{}c@{}}Lexicon\\      Test\\      PER\end{tabular}} &
  \textbf{\begin{tabular}[c]{@{}c@{}}Sentence\\      Test\\      PER\end{tabular}} &
  \textbf{\begin{tabular}[c]{@{}c@{}}Homograph\\      Test\\      PER\end{tabular}} &
  \textbf{\begin{tabular}[c]{@{}c@{}}Weighted\\ Average\\ PER\end{tabular}} &
  \textbf{\begin{tabular}[c]{@{}c@{}}Homograph\\      Accuracy\end{tabular}} \\ \hline
1  & -                         & SoundChoice                     & -                 & 9.6  & 5.7 & 2    & 6.9 & 92.2 \\ \cline{2-9} 
2  & \multirow{4}{*}{One-shot} & GPT4                            & -                 & 11.3 & 7.4 & 4.3  & 8.7 & 57.4 \\
3  &                           & Llama2-7B-chat                  & -                 & 8.4  & 3.8 & 2.1  & 5.5 & 89.2 \\
4  &                           & Gemma-2B-it                     & -                 & 8    & 3.6 & 2    & 5.2 & 83.2 \\ \cline{2-9}
5 &
  \multirow{6}{*}{\begin{tabular}[c]{@{}c@{}}LLM \\ In-context \\ Knowledge\\ Retrieval\end{tabular}} &
  \multirow{2}{*}{GPT4} &
  1 &
  9.2 &
  5.6 &
  2.4 &
  6.7 &
  95.3 \\
6  &                           &                                 & 2 & 9.2  & 3.9 & 4.5  & 6.1 & \textbf{95.7} \\ \cline{3-9} 
7 &                           & \multirow{2}{*}{Llama2-7B-chat} & 1 & 7.1  & 5.8 & 5.1  & 6.3 & 86.9 \\
8 &                           &                                 & 2 & 7.1  & 3.9 & 4.5  & 5.3 & 86.9 \\ \cline{3-9} 
9 &                           & \multirow{2}{*}{Gemma-2B-it}    & 1 & 6.1  & 6.6 & 3.7  & 6.1 & 78.7 \\
10 &                           &                                 & 2 & 6.2  & 4.0 & 5.0  & \textbf{5.0} & 79.1 \\ \hline
\end{tabular}%
}
\end{table}

\begin{table}[]
\centering
\caption{Performance in terms of PER(\%) and Homograph Accuracy(\%) for extra experiment. The dictionary 3 represent librg2p\_omit+cmu, 4 represent librg2p\_freq+cmu}
\vspace{-0.3cm}
\label{tab:addon_result}
\resizebox{\columnwidth}{!}{%
\begin{tabular}{ccccccccc}
\hline
\textbf{ID} &
  \textbf{method} &
  \textbf{model} &
  \textbf{dictionary} &
  \textbf{\begin{tabular}[c]{@{}c@{}}Lexicon\\      Test\\      PER\end{tabular}} &
  \textbf{\begin{tabular}[c]{@{}c@{}}Sentence\\      Test\\      PER\end{tabular}} &
  \textbf{\begin{tabular}[c]{@{}c@{}}Homograph\\      Test\\      PER\end{tabular}} &
  \textbf{\begin{tabular}[c]{@{}c@{}}Weighted\\ Average\\ PER\end{tabular}} &
  \textbf{\begin{tabular}[c]{@{}c@{}}Homograph\\      Accuracy\end{tabular}} \\ \hline
1 &
  \multirow{6}{*}{\begin{tabular}[c]{@{}c@{}}LLM \\ In-context \\      Knowledge \\ Retrieval\end{tabular}} &
  \multirow{2}{*}{GPT4} & 3 & 9.3  & 5.6 & 2.7  & 6.8 & \textbf{95.5} \\
4  &                           &                                 & 4 & 9.2  & 3.9 & 4.4  & 6.1 & 95.1 \\ \cline{3-9} 
5 &                           & \multirow{2}{*}{Llama2-7B-chat}  & 3 & 7.1  & 5.8 & 5.1  & 6.3 & 86.9 \\
6 &                           &                                 & 4 & 7.1  & 3.9 & 4.5  & 5.2 & 86.9 \\ \cline{3-9} 
7 &                           & \multirow{2}{*}{Gemma-2B-it}    & 3 & 6.1  & 6.6 & 3.7  & 6.1 & 78.7 \\
8 &                           &                                 & 4 & 6.2  & 4.0 & 5.0  & \textbf{4.9} & 79.1 \\ \hline
\end{tabular}%
}
\vspace{-0.4cm}
\end{table}

\section{Conclusion}
\label{section:conclusion}

To the best of our knowledge, this is the first work that leverages in-context knowledge retrieval from GPT-4 for handling the disambiguation challenge in G2P mapping process for G2P systems. In contrast, previous research has been limited to data augmentation and architectural modifications, overlooking the potential of leveraging LLMs' contextual capabilities. In this paper, in-context knowledge retrieval from LLMs was incorporated into Grapheme-to-Phoneme (G2P) conversion process to bridge the disambiguation gap in G2P mappings. Experiments on the Librig2p demonstrate that the proposed in-context knowledge retrieval-based G2P system outperforms the baseline without context information. The highest accuracy rate of 95.7\% and the lowest weighted average phoneme error rate of 4.9 \% were obtained on the Librig2p dataset compared to other non-context-based methods, which demonstrates the potential of in-context knowledge retrieval for context-aware speech synthesis applications. These empirical findings suggest that leveraging the contextual understanding capabilities of LLMs has the potential to bridge the gap in G2P mappings. As long as the semantic understanding capabilities of LLM are sufficient to distinguish the usage of words in different contexts, the system can achieve a good performance with a high-quality dictionary. How to customize a high-quality phoneme dictionary is a challenge for this system in the future.

\section{Acknowledgement}
This research is supported by the National Natural Science Foundation of China (62306260) and the Centre for Perceptual and Interactive Intelligence.

\bibliographystyle{IEEEtran}

\bibliography{main}


\end{document}